\pdfoutput=1

\documentclass[11pt]{article}

\usepackage[]{ACL2023}

\usepackage{times}
\usepackage{latexsym}
\usepackage{amsfonts}

\usepackage{multirow}
\usepackage{multicol}
\usepackage{booktabs}
\usepackage{tabularx}
\usepackage{graphicx}
\usepackage{float}
\usepackage{amsmath}
\usepackage{xcolor}
\usepackage{makecell}

\usepackage[T1]{fontenc}

\usepackage[utf8]{inputenc}

\usepackage{microtype}

\usepackage{inconsolata}

\usepackage{array,booktabs}

\title{Attend or Perish: Benchmarking Attention in Algorithmic Reasoning}

\author{Michal Spiegel$^{\clubsuit \diamondsuit}$\thanks{\hspace{4pt}Correspondence to: {spiegel.michal@gmail.com}} \and Michal Štefánik$^{\heartsuit \clubsuit}$ \and Marek Kadlčík$^\clubsuit$ \and Josef Kuchař$^\clubsuit$ \vspace{6pt}\\
  \normalsize{$^\clubsuit$TransformersClub @ Faculty of Informatics, Masaryk University} \\
  \normalsize{$^\diamondsuit$Kempelen Institute of Intelligent Technologies} \\
  \normalsize{$^\heartsuit$Language Technology, University of Helsinki} 
 }

\begin{document}
\maketitle
\begin{abstract}
Can transformers learn to perform algorithmic tasks reliably across previously unseen input/output domains?
While pre-trained language models show solid accuracy on benchmarks incorporating algorithmic reasoning, assessing the reliability of these results necessitates an ability to distinguish genuine algorithmic understanding from memorization.
In this paper, we propose AttentionSpan, an algorithmic benchmark comprising five tasks of infinite input domains where we can disentangle and \textit{trace} the correct, robust algorithm necessary for the task.
This allows us to assess (i)~models' ability to \textit{extrapolate} to unseen \textit{types} of inputs, including new lengths, value ranges or input domains, but also (ii)~to \textit{assess} the robustness of their learned mechanisms. By analyzing attention maps and performing targeted interventions, we show that attention mechanism directly causes failures in extrapolation.
We make the implementation of all our tasks and interpretability methods publicly available.\footnote{\scriptsize{\url{https://github.com/michalspiegel/AttentionSpan}}
}
\end{abstract}

\section{Introduction}

The neural architecture of Transformer~\cite{attention-all-you-need} presents a backbone for a vast majority of modern language processing applications.
A growing body of these applications, including code generation, conversational assistants, or data processing automation, requires Transformers to exhibit robust \textit{reasoning}, i.e., an ability to identify and combine relevant pieces of information to infer \textit{new} information~\cite{yu2023reasoning}. 

Transformers rely on the Attention mechanism~\cite{chov1} as their only means of mixing information across token streams, which is essential for long-form reasoning. However, its effectiveness often degrades, especially with increasing sequence lengths~\cite{velickovic2025softmax}.
Despite the strong theoretical expressivity~\cite{Yun2020-universal-approximators,merrill2024the}, Transformers often depend on spurious features of data~\cite{mikula2023think}, causing even high-end models to fail in unexpected scenarios. 
This unreliability currently presents a critical bottleneck across a variety of applications. 
Bridging this gap requires fundamental improvements not only in \textit{architecture} ~\cite{ye2025differential,velickovic2025softmax} but also \textit{evaluation} to rigorously assess how \textit{robust} the reasoning process of our models is.

In this work, we contribute to bridging this gap by creating AttentionSpan, a new evaluation suite focused on assessing fundamental reasoning capabilities of language models in out-of-distribution scenarios.

\begin{figure}[t]
\centering
  \!\!\!\!\includegraphics[width=1.03\columnwidth]{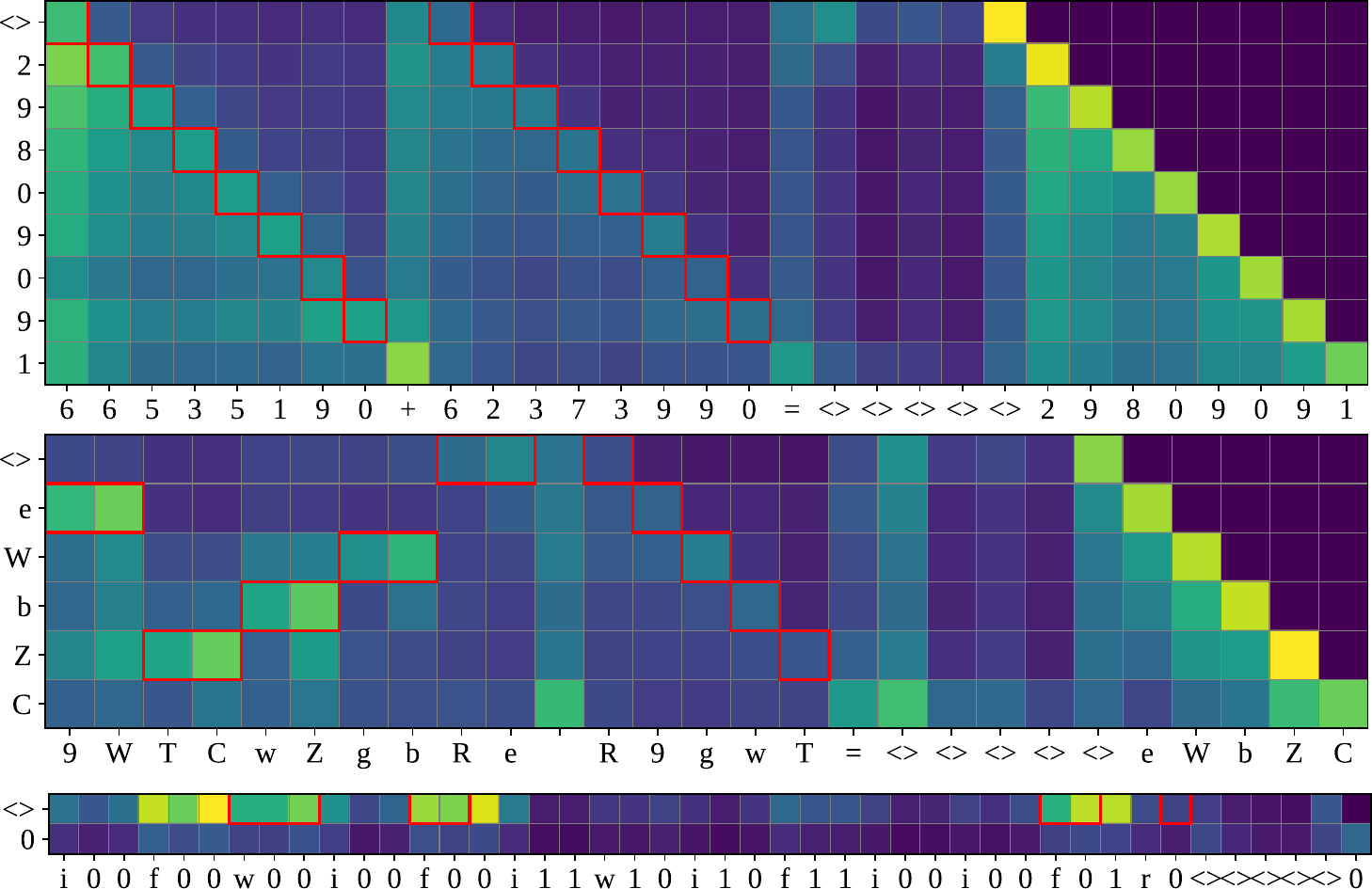}
  \caption{Examples of attention patterns that we use to evaluate the robustness of models' reasoning. Each grid displays normalized attention scores (post-softmax), aggregated across heads and layers via attention rollout, overlaid with a \textit{reference map} (in red) denoting necessary input tokens. Visualizations show the tasks of addition, value assignment, and FFML (§\ref{sec:attentionspan}; ordered top to bottom, respectively). 
  }
  \label{fig:example_attention_maps}
\end{figure}

Each task in AttentionSpan has a solver algorithm, which generates a step-by-step solution, together with a trace of \textit{which} past tokens are necessary for correctly generating the next one. 
This enables deeper analyses of the model's reasoning behavior, not possible with previous resources: 
\begin{itemize}
    \item Provision of \textbf{reference attention masks} representing the ground-truth reasoning patterns that a successful model has to follow in order to achieve a correct prediction.
    \item Control over \textbf{distribution shifts}, i.e., systematic changes in the constructed dataset that allow for a reliable assessment of models' reasoning in novel scenarios.
\end{itemize}

First, we apply our benchmark to evaluate two different facets of generalization of existing language models: (i)~an ability to learn to accurately \textit{combine} the information necessary for the task in-distribution, uncovering the models ability to fit an algorithm, and (ii)~an ability to \textit{generalize} to out-of-distribution data, exploiting the inherent limitations of existing architectures;
We find that Transformer models can learn to explain in-distribution (ID) data by a seemingly correct algorithm, but despite that, fail to generalize to unseen, out-of-distribution (OOD) inputs.

Finally, we hypothesize that out-of-distribution (OOD) errors stem from attention misalignments and verify this by an intervention experiment that directly manipulates attention to match reference attention maps from AttentionSpan. We find that such intervention \textit{increases} models' OOD accuracy by up to 90 percentage points, and provides \textit{consistent} improvements even with increasing input sequence lengths. This establishes a direct causal link between attention failures and poor generalization, pinpointing attention as a key bottleneck for length extrapolation.


The presented benchmark empowers future work by a toolset for further, deeper analyses of models' internal functioning, isolated from other confounders such as memorization.
Our findings also motivate future work in architectural refinements, particularly those addressing limitations of the current Attention mechanism.


\section{Related Work}

Closely related to our work, CRLS-Text \cite{markeeva2024clrstextalgorithmicreasoninglanguage} is a benchmark for algorithmic reasoning that implements traditional algorithms to train and evaluate state-of-the-art LLMs. We build upon the methodology of CRLS-Text and extend it to allow for, not only accessing the performance, but also to provide means for interpretation and investigation of the results by means of the reference attention masks.

BIG-Bench \cite{srivastava2023imitationgamequantifyingextrapolating} is a massive benchmark comprised of more than 200 tasks, many of which specialize in evaluating algorithmic reasoning, e.g. addition or dyck languages. However, as a fixed test set, it is hard to use it to robustly evaluate models on extrapolation, while the recent work finds that BIG-Bench was indeed leaked into the training data of recent models \cite{fajcik2024benczechmark}, including Qwen. We extend the tasks from BIG-Bench into configurable generators capable of generating infinite data, enabling a controlled training and evaluation setting, avoiding data contamination.

Flip-Flop Language Modeling is a synthetic task introduced by \citet{liu2023exposingattentionglitchesflipflop}. The authors introduce this simple algorithmic task to analyze hallucinations caused by attention glitches. We extend this idea and implement novel analysis of attention on a number of diverse algorithmic tasks. 

\section{AttentionSpan: Dataset and Evaluation Suite}
\label{sec:attentionspan}

To evaluate the reasoning robustness of Transformers, we introduce AttentionSpan, a benchmark for analyzing models' attention patterns in step-by-step reasoning tasks. 

AttentionSpan is composed of synthetic tasks allowing for a fully controlled generation setting and guaranteed out-of-distribution assessment. The tasks include string reversal, addition, multiplication, flip-flop language modeling~\cite{liu2023exposingattentionglitchesflipflop}, and value assignment, all described in detail in Appendix~\ref{sec:task_description}. Examples of inputs and outputs can be found in Table~\ref{table:problem-examples}. 
Task instances (problems) can be randomly generated in arbitrary quantities and with configurable difficulty. 
The configuration also allows for systematic ID-OOD splits applied in our evaluations, including input lengths, ranges or domains.
We detail our evaluated ID-OOD configurations of AttentionSpan's tasks in Appendix~\ref{sec:ood_evaluation}.

\begin{table*}[tbh]
    \renewcommand{\arraystretch}{1.2}
    \centering
    \scalebox{0.8}{
    \begin{tabular}{|l|l|l|}
    \hline
    \textbf{Task}       & \textbf{Example Input}                    & \textbf{Corresponding Output}                  \\
    \hline
    String Reversal     & d h 1 3 h 8 2 h j 2 8 3 j 2 3 H = & H 3 2 j 3 8 2 j h 2 8 h 3 1 h d  \\
    Long Addition       & 1240 + 4335 + 3440 =              & 8916                             \\
    Long Multiplication & 9900 * 9900 =                     & 1980 + 0198 + 0000 + 0000 = 1089 \\
    FFLM                & w 1 1 i 1 1 f 1 0 r 1 0 f 1 0 r 1 & 1                                \\
    Value Assignment    & B1 E0 D1 A1 C0 \, ABBEDACABCD           & 11101101101                      \\
    \hline
    \end{tabular}
    }
    \caption{
        Example instances of our tasks. The spacing is adjusted for clarity and does not denote a separator of tokens. How the tasks handle tokenization is described in greater detail in Appendix~\ref{sec:tokenization}
    }
    \label{table:problem-examples}
\end{table*}

\subsection{Reference Attention Masks}

%
%
%

A key novelty of AttentionSpan is the provision of a \textit{reference attention mask} with every generated data sample. This mask precisely identifies the past tokens essential for correct next-token inference. We utilize these masks as an expected attention pattern for an ideal model and subsequently measure the alignment of a model's learned attention with this reference. Our experiments demonstrate that reference attention masks are a powerful tool for diagnosing reasoning errors in Transformers, evidencing that they could facilitate future research aimed at enhancing model reliability, e.g., through architectural modifications.

The reference attention mask is a discrete boolean matrix. For each target token to be predicted, it identifies critical (reference) past tokens that the model should attend to for robust problem-solving (see tokens highlighted in red in Figure~\ref{fig:example_attention_maps}). An element in the matrix is set to 1 if the corresponding past token carries information relevant to predicting the target token; otherwise, it is set to 0, indicating irrelevant tokens that may not be attended for the current prediction. 


\begin{table}[t]
    \centering
    \scalebox{0.7}{
    \begin{tabularx}{1.3\columnwidth}{p{0.8cm}l XXX}
        \toprule
        \textbf{Model} & \textbf{Task} & \textbf{ID Acc.} & \textbf{OOD Acc.} & \textbf{OOD Partial Acc.} \\
        \midrule
        \multirow{5}{*}{\rotatebox{90}{\textit{Llama-3.2}}}
           & String Reversal       & 95.83 & 53.83 & 96.18 \\
           & Long Addition         & 96.87 & 1.61 & 64.76\\
           & Long Multiplication   & 86.00 & 0.00 & 73.87\\
           & FFML                  & 100.00 & 99.20 & 99.79\\
           & Value Assignment      & 93.95 & 0.52 & 65.26\\
        \midrule
        \multirow{5}{*}{\rotatebox{90}{\textit{Qwen2.5}}}
           & String Reversal       & 98.95 & 21.77 & 76.26 \\
           & Long Addition         & 100.00 & 44.75 & 92.77 \\
           & Long Multiplication   & 56.25 &  0.00 & 80.14 \\
           & FFML                  & 100.00 & 88.50 & 96.71\\
           & Value Assignment      & 76.04 & 1.04 & 81.06\\
        \midrule
        \multirow{5}{*}{\rotatebox{90}{gemma-3}}
           & String Reversal       & 96.87 & 6.04 & 37.64 \\
           & Long Addition         & 100.00 & 2.62 & 67.85\\
           & Long Multiplication   & 89.58 & 0.00 & 77.16\\
           & FFML                  & 100.00 & 90.12 & 96.71\\
           & Value Assignment      & 98.95 & 0.00 & 19.35\\
        \bottomrule
    \end{tabularx}}
    \caption{Accuracy of finetuned models on AttentionSpan tasks with consistent in-distribution and out-of-distribution splits. Despite a sharp decline in OOD Accuracy in almost all cases, the OOD Partial Accuracy reveals that models correctly predict a large proportion of target tokens, indicating some extrapolation abilities are present.
    \vspace{-5pt}
    }
    \label{tab:overall_performance}
\end{table}

\begin{table}[t]
    \centering
    \scalebox{0.7}{
    \begin{tabularx}{1.3\columnwidth}{p{0.5cm} l c c c c}
        \toprule
        \textbf{Model} & \hspace{2em} \textbf{Task} & \textbf{\makecell[c]{Mean AttnScore \\ (Correct) (\%)}} & \textbf{\makecell[c]{Mean AttnScore \\ (Error) (\%)}} \\
        \midrule
        \multirow{2}{*}{\rotatebox{90}{\scalebox{0.6}{\textit{Llama-3.2}}}}
           & String Reversal       & $4.55 \pm 0.02$ & $2.36 \pm 0.09$ \\
           & Value Assignment      & $3.33 \pm 0.03$ & $1.26 \pm 0.02$ \\
        \midrule
        \multirow{2}{*}{\rotatebox{90}{\scalebox{0.6}{\textit{Qwen2.5}}}}
           & String Reversal       & $3.07 \pm 0.06$ & $2.17 \pm 0.05$ \\
           & Value Assignment      & $1.27 \pm 0.03$ & $1.20 \pm 0.06$ \\
        \midrule
        \multirow{2}{*}{\rotatebox{90}{\scalebox{0.6}{\textit{gemma-3}}}}
           & String Reversal       & $3.97 \pm 0.12$ & $2.30 \pm 0.03$  \\
           & Value Assignment      & $5.04 \pm 0.12$ & $4.91 \pm 0.04$ \\
        \bottomrule
    \end{tabularx}}
    \caption{\textbf{Errors in prediction are associated with lower attention score.} We find a statistically significant difference (Welch's t-test) between attention scores on correct and incorrect (target token) predictions.
    }
    \label{tab:broken_attention}
\end{table}

\section{Experiments and Evaluation}
\label{sec:experiments}

\begin{figure*}[th!]
\centering
  \includegraphics[width=0.92\textwidth]{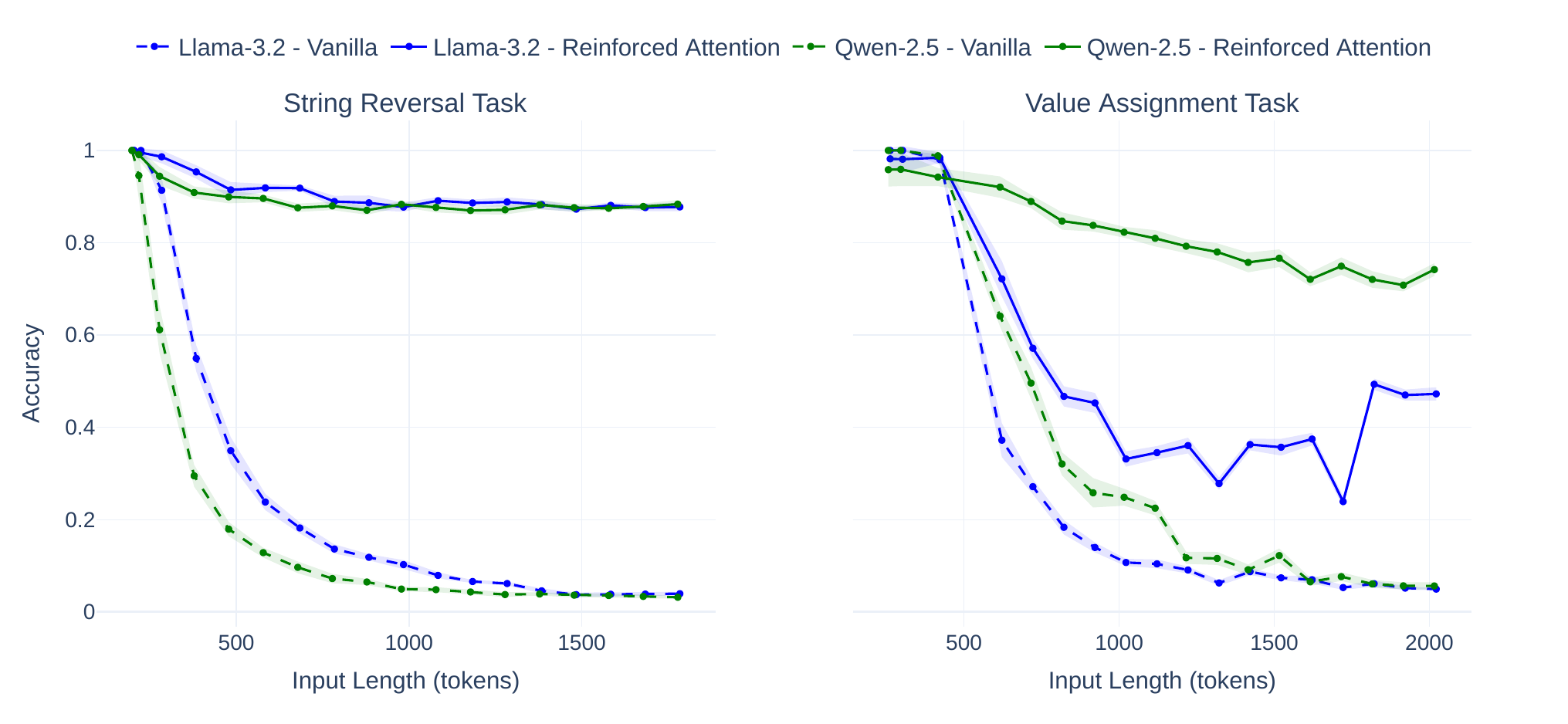}
  \caption{We demonstrate that by intervening on models' (\textit{Llama-3.2-1B-Instruct}, \textit{Qwen2.5-1.5B-Instruct}) activations and directly adjusting their attention scores to reinforce our reference attention pattern, we are able to drastically improve the length extrapolation performance over the vanilla models without invervention, attributing the failure to extrapolate to the attention mechanism. }
  \label{fig:attention_reinforcement}
\end{figure*}

Building on AttentionSpan, we assess how robustly recent language models can learn and execute the underlying algorithm. Towards this goal, we explore models in two distinct settings: (1) training models from random initialization, and (2) fine-tuning pre-trained models -- specifically \textit{Llama-3.2-1B-Instruct}~\cite{grattafiori2024llama3herdmodels}, \textit{Qwen2.5-1.5B-Instruct}~\cite{yang2024qwen2technicalreport}, and \textit{gemma-3-1b-it}~\cite{gemmateam2025gemma3technicalreport} -- on all tasks using a few-shot instruction prompt, in order to evaluate the contribution of pre-training. To evaluate these models, we use AttentionSpan to introduce systematic distribution shifts, altering the length of input sequences and other key parameters (for details on ID/OOD settings see Appendix~\ref{sec:ood_evaluation}). Training setup and hyperparameter search are further detailed in Appendix~\ref{sec:training_hyperparam}. 

In addition to strict exact-match accuracy of whole output sequences, we measure \textbf{Partial Accuracy} as the fraction of correct next-token predictions in the target sequence:

\begin{equation}
\text{Partial Acc} = \frac{1}{N} \sum_{i=1}^{N} \frac{1}{|T_i|} \sum_{j=1}^{|T_i|} \mathbb{I}(\hat{y}_{i,j} = y_{i,j})
\label{eq:partial_acc}
\end{equation}
where $N$ is the total number of samples, $T_i$ is the target sequence for sample $i$, $\hat{y}_{i,j}$ and $y_{i,j}$ are the predicted and target tokens at position $j$ respectively, and $\mathbb{I}(\cdot)$ is the indicator function.

We use the reference attention masks to track which tokens the model considers at each reasoning step. We rely on \textbf{attention rollout}~\cite{abnar2020quantifyingattentionflowtransformers}, a standard method for aggregating attention across heads and layers. This allows us to visualize the attention patterns (e.g., Figure~\ref{fig:example_attention_maps}) and assess whether the model learned the expected attention patterns in a generalized fashion. The formula for attention rollout (as defined in \cite{abnar2020quantifyingattentionflowtransformers}) can be expressed as a recursive product:
\begin{equation}
R = \prod_{l=1}^{L} \left( \frac{1}{H} \sum_{h=1}^{H} A_{l,h} + I \right)
\label{eq:rollout}
\end{equation}
where $R$ is the final attention rollout matrix, $L$ is the total number of self-attention layers, $H$ is the number of attention heads in each layer, $A_{l,h}$ is the (post-softmax) attention score matrix for head $h$ in layer $l$, $I$ is the identity matrix, which accounts for the residual connections.

Quantitatively, we leverage our dataset's reference attention masks and aggregated attention scores to calculate the proportion of attention scores assigned to tokens identified as essential for correct prediction, which we refer to as \textbf{Attn Score} in Table~\ref{tab:broken_attention} (see Appendix~\ref{sec:attention_score} for details). Subsequently, through the lens of Attn Score, we explore differences in models' behavior on correct and incorrect (token-level) predictions to identify systematic attention patterns associated with errors. Formally, Attn Score is defined as:
\begin{equation}
\text{Attn Score} \!=\! \frac{1}{N} \sum_{i=1}^{N} \!\!\left( \!\frac{1}{|T_i|} \sum_{j=1}^{|T_i|} \!\left( \sum_{k \in S_i} a_{i,j,k} \!\right) \!\!\right)
\label{eq:attn_score}
\end{equation}
where $N$ is the total number of samples in the dataset, $T_i$ is the sequence of target (output) tokens for sample $i$, and $|T_i|$ is its length, $S_i$ is the set of indices corresponding to the reference input tokens for sample $i$, $a_{i,j,k}$ is the normalized attention score in aggregated $R$ (Eq.~\eqref{eq:rollout}) from the $j$-th output token to the $k$-th input token in sample $i$. We will represent AttnScore as percentages.

\section{Results: Attention as the Bottleneck}

As detailed in Table~\ref{tab:overall_performance}, the fine-tuned models exhibit difficulty generalizing to OOD data (See Appendix~\ref{sec:ood_evaluation} for details on exact ID/OOD parameters). However, their ability to correctly predict a majority of target tokens in each sample (demonstrated by high partial accuracy) suggests underlying length extrapolation capabilities and significant potential for future improvement.

Models trained from random initialization, despite fitting the in-distribution (ID) data, exhibited near-zero accuracy in OOD evaluations (see Appendix~\ref{sec:train_random_init}). Given this lack of generalization, we focused our subsequent analysis on fine-tuned models, with the full results from this initial experiment available in Appendix~\ref{sec:train_random_init}. This outcome suggests that the pre-training phase is advantageous for generalization and convergence.

Our analyses reveal that for String Reversal and Value Assignment, out-of-distribution (OOD) errors are linked to reduced attention on reference tokens (Table \ref{tab:broken_attention}). The appearance of this phenomenon in these specific tasks is significant, as they are representative of copying tasks—a recently highlighted benchmark class \cite{arjovsky2016unitaryevolutionrecurrentneural, jelassi2024repeatmetransformersbetter}—and embody two distinct, fundamental retrieval modes: positional (e.g., sorting, reordering) and content-based (e.g., using variables or a knowledge base). This pattern suggests that insufficient attention to reference tokens contributes to faulty predictions. We observe this issue across diverse pre-trained models and architectures, indicating a naturally emergent, general problem.

\subsection{Attention Misallocation as a Failure Mode}

\begin{figure}[t]
  \centering
  \scalebox{0.9}{
  \includegraphics[width=\columnwidth]{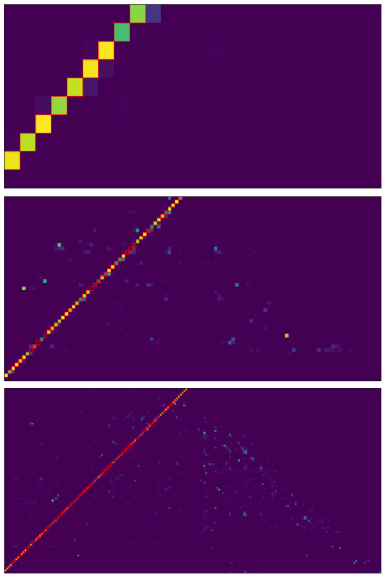}}
  \caption{Attention from a single head for String Reversal identified as performing our reference diagonal lookup algorithm. As sequence length increases from in-distribution (top) to out-of-distribution (bottom), attention scatters from the correct diagonal. This misallocation (highlighted in red) directly causes prediction errors.}

  \label{fig:string_reversal_attention_scatter}
\end{figure}

To better understand the nature of these OOD failures, we visualize the attention patterns of individual heads that specialize in the required lookup algorithm. Figure~\ref{fig:string_reversal_attention_scatter} demonstrates how attention scores scatter away from the reference diagonal as input sequence length increases. Table~\ref{tab:broken_attention} quantifies this, showing that prediction errors occur precisely when attention deviates from the reference diagonal. Interestingly, we observe that attention does not simply diffuse across neighboring tokens in OOD scenarios. Instead, it remains sharp but incorrectly shifts to a distant, irrelevant token. This suggests a potential failure in how positional embeddings like RoPE generalize to long-range positional relationships not seen during training.

\subsection{Attention Reinforcement}

To causally validate that insufficient attention to reference tokens drives OOD failures, we first identified key attention heads implementing our expected reference attention pattern. To do this, we select the top 10\% of the heads with the largest sum of attention scores on \textit{reference} tokens in ID setting. We then selectively reinforce the reference attention in OOD inference by directly increasing the post-softmax attention to reference tokens of such-identified heads. As shown in Figure~\ref{fig:attention_reinforcement}, this intervention leads to an absolute increase in accuracy on OOD samples of up to 90 percentage points, consistently across all input lengths. This provides \textbf{causal evidence that insufficient attention to reference tokens largely contributes to extrapolation failures}. For details on the implementation of the intervention, see Appendix~\ref{sec:attn_inter_details}.

While our intervention targets attention outputs, we hypothesize the root cause is the extrapolation failure of positional embeddings like RoPE. The key-query similarity function is learned over a limited range of relative distances. For longer OOD sequences, the model encounters distances outside this familiar range, causing the learned function to fail and produce the unstable, misallocated attention scores we observe.

\section{Conclusion}

We introduced AttentionSpan, a novel algorithmic reasoning benchmark including reference attention masks, allowing assess Transformer extrapolation under different, parametrized distribution shifts, thus minimizing memorization effects. Our evaluations reveal that while pre-training helps, models struggle with out-of-distribution (OOD) generalization even on simple algorithmic tasks. Through attention analysis and targeted interventions, we causally attribute these OOD failures to Attention's inability to robustly identify essential tokens. We show that reinforcing correct attention patterns significantly improved OOD accuracy and length extrapolation, pinpointing attention as a critical bottleneck.

\section*{Limitations}
We identify several limitations of our work and mention what we believe are the main ones below.
First, our interpretability of models' internal functioning builds upon the assumption that models robustly executing the correct algorithm should fully attend only to tokens that are relevant to the algorithm.
Nevertheless, we note that even a model with a systematic dispersion of attention across irrelevant tokens might still be able to robustly execute algorithm, as long as the irrelevant attended tokens do not significantly alter the attention's output representations.
Therefore, there is not a necessary equivalence between the model's robustness and accuracy of attention with respect to our references. 
However, in the situation where the model does not attend the relevant tokens at all, we can still claim that the model does not represent the task's correct/robust algorithm.

Finally, we note the limitation in using a single interpretability method in our analyses in Section~\ref{sec:experiments} (Attention rollout).
While we argue that this method best represents the computation flow within the transformer across tokens, it still does not take into account some computation parts of the model, such as the impact of feed-forward layers which might, theoretically, exclude the impact of even some attended tokens.


\bibliography{custom,stefanik}
\bibliographystyle{acl_natbib}

\appendix

\section{Task Descriptions}
\label{sec:task_description}

\subsection{String Reversal}

This task requires the model to generate the input sequence in the reverse order. The task generator can be configured by the character set and the range of the input length.

\subsection{Long Multiplication}

Long multiplication is parametrized by the digit length of two operands and optional padding. The solution contains a sequence of intermediate products, which are then summed together into the final result. The digit ordering is consistent with the long addition task.

\subsection{Long Addition}

This task consists of adding several multi-digit numbers. The digits are ordered from the least significant to the most significant. The ordering of the digits is given by the standard addition algorithm where we compute the lower order digits first in order to be able to propagate the carry to the topmost digit. The problem generator can be parametrized by the number of operands, their length in digits, and whether short numbers are padded with zeros. As a subtask of long multiplication, it provides further insight into the inner functioning of models on these arithmetic tasks.

\subsection{Value Assignment}

In this task, the problem specifies a translation table from an input alphabet to an output alphabet. The model is then required to translate an input string, symbol by symbol. The character sets, and the string length can be configured. Value assignment is a subtask of many algorithmic tasks where we work with symbolic representations.

\subsection{Flip Flop Language Modeling}

Flip Flop Language Modeling, as introduced by \cite{liu2023exposingattentionglitchesflipflop} represents a simulation of memory composed of a single one-bit registers. We extend this into multiple registers problem, adding a new flip command that flips the value of the specific register. The input is a sequence of read, write, ignore, and flip instructions, each with the register index specified as a first operand. The sequence ends with a read instruction, and the solution is the bit value currently stored at the selected register. The parameters of the task can specify how many registers are used, the length of the instruction sequence, and whether flip commands are used. 

\section{Attention Score on Reference Tokens}
\label{sec:attention_score}
The proportion attention score attributed to reference tokens is computed per each row of the aggregated attention, that is for each predicted token, separately. This attributes to the need to investigate the proportion of information that has influenced a given output representation or output token. The result is then averaged across the whole sample or the whole batch to get an idea of how the model attributes attentions score on a given distribution of data.

\section{Tokenization of training and evaluation samples}
\label{sec:tokenization}

With the exclusion of the instruction prompt, we tokenize the few-shot examples and the data points themselves into single character-level tokens. This is important to prepare the reference attention masks. Without tokenizing like this it would be possible to evaluate the attention patterns because different tokenization schemes wildly change the nature of the task and distribution of critical information between tokens. However, the fine-tuned models were able to parse this representation and fit the task as can be seen in the resulting accuracies after training.

\section{Training Hyperparameters}
\label{sec:training_hyperparam}
The following configuration summarizes the setup used for fine-tuning (or training from scratch) of our models.

\textbf{Model:}
\begin{itemize}
  \item \textbf{Name:} meta-llama/Llama-3.2-1B-Instruct
  \item \textbf{Architecture Configuration:}
    \begin{itemize}
      \item Attention Dropout Probability: 0.0
      \item Hidden Dropout Probability: 0.0
    \end{itemize}
\end{itemize}

\textbf{Training Hyperparameters:}
\begin{itemize}
  \item \textbf{Epochs:} 1
  \item \textbf{Batch Size:} 4
  \item \textbf{Optimizer:} AdamW
  \item \textbf{Optimizer Parameters:}
    \begin{itemize}
      \item Learning Rate: $5\times10^{-6}$
      \item $\beta_1$: 0.95
      \item $\beta_2$: 0.999
      \item Weight Decay: 0.2
    \end{itemize}
\end{itemize}

These hyperparameters are chosen on the basis of a hyperparameter search that was executed on String Reversal and Addition tasks, the results of the search was averaged over these two tasks. The hyperparameter search can be reproduced by running the prepared script in our codebase.

The conclusion of the hyperparameter search was that, for both tasks, smaller batch size, smaller learning and weight decay were effective in increasing accuracy in OOD. The effect of using dropout in attention or hidden layers was highly task-dependent and inconclusive, so we decided not to use it. 

All our experiments were run on a single Nvidia A100 GPU card and required less than 12 hours to converge.
As we document in our codebase, our experiments employ HuggingFace Transformers library~\cite{Wolf2019HuggingFacesTS} v4.48.1 and PyTorch v2.5.1.

\section{OOD Evaluation}
\label{sec:ood_evaluation}

\subsection{Long Addition Task Evaluation Parameters}

The following configuration details the evaluation setup for the Long Addition task.

\textit{In-distribution:}
\begin{itemize}
  \item 2 operands
  \item Each number is 1-4 digits long
\end{itemize}

\textit{Out-of-distribution:}
\begin{itemize}
  \item 2 operands
  \item Each number is 5-10 digits long
\end{itemize}

\subsection{FFML Task Evaluation Parameters}

The following configuration details the evaluation setup for the FFML task.

\textit{In-distribution:}
\begin{itemize}
  \item Use the flip command
  \item Each string is composed of 10 commands
  \item Each instance works with 2 different registers
\end{itemize}

\textit{Out-of-distribution:}
\begin{itemize}
  \item Use the flip command
  \item Each string is composed of 11-100 commands
  \item Each instance works with 2 different registers
\end{itemize}

\subsection{Long Multiplication Task Evaluation Parameters}

The following configuration details the evaluation setup for the Long Multiplication task.

\textit{In-distribution:}
\begin{itemize}
  \item Each number is 1-3 digits long
\end{itemize}

\textit{Out-of-distribution:}
\begin{itemize}
  \item Each number is 4-6 digits long
\end{itemize}

\subsection{String Reversal Task Evaluation Parameters}

The following configuration details the evaluation setup for the String Reversal task.

\textit{In-distribution:}
\begin{itemize}
  \item Each string is 1-10 characters long
  \item The character set is composed of at least 50 unique characters
\end{itemize}

\textit{Out-of-distribution:}
\begin{itemize}
  \item Each string is 11-50 characters long
  \item The character set is composed of at least 50 unique characters
\end{itemize}

\subsection{Successor Task Evaluation Parameters}

The following configuration details the evaluation setup for the Successor task.

\textit{In-distribution:}
\begin{itemize}
  \item The starting number is between 1 and 90
  \item The length of the series is 2-4 numbers
\end{itemize}

\textit{Out-of-distribution:}
\begin{itemize}
    \item The starting number is between 100 and 900
    \item The length of the series is 5-6 numbers
\end{itemize}

\subsection{Value Assignment Evaluation Parameters}

The following configuration details the evaluation setup for the Value Assignment task.

\textit{In-distribution:}
\begin{itemize}
  \item The number of unique tuples in the translation table is 5
  \item The length of the string to be translated is 5
\end{itemize}

\textit{Out-of-distribution:}
\begin{itemize}
  \item The number of unique tuples in the translation table is 10-50
  \item The length of the string to be translated is 10-20
\end{itemize}

\section{Attention Intervention Details}
\label{sec:attn_inter_details}

Our intervention method aims to causally link attention deficits to out-of-distribution (OOD) performance degradation by selectively reinforcing attention to reference tokens. The process involves two main stages: identifying key attention heads and applying the intervention.

\begin{enumerate}
    \item \textbf{Identifying Key Attention Heads}: To pinpoint the attention heads most responsible for implementing the desired reference attention pattern, we perform the following steps:

    \begin{itemize}
        \item We run inference on multiple in-distribution (ID) data samples.
        \item For each attention head, we calculate the sum of its post-softmax attention scores on the pre-defined reference tokens. This sum is accumulated across all ID samples.
        \item This cumulative scoring helps identify heads that consistently attend to reference tokens, as well as those that might activate for specific patterns present only in a subset of samples (e.g., particular carry operations in addition tasks).
        \item Heads are then ranked in descending order based on this cumulative score.
        \item We select the top N heads for intervention where N is a hyperparameter we optimize to achieve the largest performance improvement on the end-to-end task (e.g., string reversal) post-intervention.
    \end{itemize}
    \item \textbf{Applying the Intervention during OOD Inference}: The intervention is applied exclusively to the N selected heads during OOD inference.
    \begin{itemize}
        \item \textbf{Standard Intervention} (String Reversal): For each selected head, we directly modify its post-softmax attention scores. A constant value C (a hyperparameter, typically ranging from 0.3 to 2.0) is added to the attention score of every token position corresponding to a reference token. These modified attention scores are then propagated through the network. This approach proved effective for tasks like string reversal.
        \item 
        \textbf{Conditional Intervention} (e.g., for Value Assignment): For more complex tasks like value assignment, we observed that the reference attention pattern was often distributed across multiple heads, and a simple global reinforcement was ineffective. Instead, we adopted a conditional reinforcement strategy:
        \begin{itemize}
            \item For each selected head, we add the constant C to the post-softmax attention score at a reference token position only if the original attention score at that specific position already exceeds a certain threshold (another optimizable hyperparameter).
             \item This approach reinforces existing, albeit potentially weak, attention sub-patterns within a head, rather than imposing the entire reference pattern uniformly.
            \item The conditional intervention for value assignment, while improving performance, sometimes results in a slightly lower accuracy boost compared to the standard intervention on simpler tasks. This is because if the initial activation for a crucial reference token falls below the threshold, our intervention, by design, will not reinforce it, even if doing so would be beneficial.
        \end{itemize}

    \end{itemize}
    
\end{enumerate}

\section{Training models from random initialization}
\label{sec:train_random_init}

\begin{table}[t]
    \centering
    \scalebox{0.7}{
    \begin{tabularx}{1.3\columnwidth}{p{0.8cm}l XX}
        \toprule
        \textbf{Model} & \textbf{Task} & \textbf{ID Acc.} & \textbf{OOD Acc.} \\
        \midrule
        \multirow{6}{*}{\rotatebox{90}{From Scratch}} 
           & String Reversal       & 5.21 & 0.0578 \\
           & Long Addition         & 9.37 & 0.1713 \\
           & Long Multiplication   & 18 & 0.1302 \\
           & FFML                  & 68.75 & 0.0129 \\
           & Value Assignment      & 4.17 & 0.3060 \\
           & Successor             & 100 & 0.4069 \\
        \bottomrule
    \end{tabularx}}
    \caption{Performance of models trained from random initialization. As mentioned in the main paper, we initiated experiments training models from scratch to evaluate performance without the benefit of pre-training. The results show that while the models could achieve some accuracy on the in-distribution (ID) data, they consistently failed to generalize, with out-of-distribution (OOD) accuracy remaining near-zero across all tasks. Due to this poor generalization performance, we did not pursue this line of research further.
    \vspace{-5pt}
    }
    \label{tab:overall_performance}
\end{table}

\end{document}